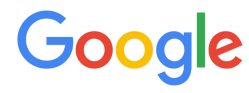

# Multi-Modal Multi-Agent Robotic Cognitive Alignment enabled by Non-Invasive Consumer Brain Computer Interfaces: A Proof of Concept Exploration

*Pre-print. Under review.*

NATALIYA KOSMYNA *

GOOGLE, Paradigms of Intelligence, Cambridge, MA, United States, kosmyna@google.com

LIZ JENKINS

GOOGLE, Paradigms of Intelligence, Mountain View, CA, United States, lizjenkins@google.com

ANOOP K. SINHA

GOOGLE, Paradigms of Intelligence, Mountain View, CA, United States, anoopsinha@google.com

While non-verbal behaviors and expressive movements are essential for natural human-robot interaction, existing methods often overlook a crucial element: the human's internal cognitive state. Frequently, proactive multi-agent systems can interrupt humans at inopportune moments, leading to cognitive overload and decreased task performance. This paper introduces a framework for generating "cognitively aligned" multi-agent interactions, enhancing the ability of robotic systems to contextually defer communications to the user of an agent system during moments of high human mental workload and engagement. We present the design and implementation of a closed-loop architecture that explores the interplay between autonomous task execution and real-time neurophysiological focus. Using a consumer-grade Brain-Computer Interface (BCI), our approach continuously monitors Electroencephalography (EEG) spectral band powers while a human performs an engagement-inducing task. We propose an engagement-driven pipeline where an HTTP-based signaling mechanism places a primary agent's sensory inputs and audio outputs into a holding state upon detecting high engagement. This allows secondary agents to seamlessly process complex, delegated tasks in the background. Once the human's cognitive state returns to a lower cognitive load baseline, the primary agent releases the queued agent message. Our preliminary results demonstrate the feasibility of leveraging real-time signal processing, Large Language Models (LLMs), and physical robotic embodiments to create cognitively-aware, non-intrusive multi-agent systems.

**Video Link**

Figure 1: Top part – 2008 version of Google Chrome with a t-rex dinosaur "Stan" in the offline game. Bottom part – 2026 version of embodied robotic agents in the shape of t-rex dinosaur "Nadine", as well as "Red Claw" and "Blue Claw" lobsters.

* Corresponding author, all inquires are to be directed to Dr. Kosmyna's email

## 1. INTRODUCTION

The rapid proliferation of Large Language Models (LLMs) and advanced Voice Activity Detection (VAD) systems has accelerated the deployment of proactive [1], multi-agent [2] systems in daily environments. However, a fundamental asymmetry exists in how humans and artificial agents operate. Humans process things differently using executive functioning; sustaining focus on a complex task requires deep, continuous cognitive immersion. In contrast, multi-agent systems are inherently parallel, capable of processing massive amounts of data, executing background API calls, and generating asynchronous outputs at high speeds and high bandwidth.

How might humans and agents communicate effectively, given to differences in perception, reasoning, action and timing of actions? An autonomous agent that interjects with an output precisely when a human is deeply focused on the task at hand, might force an abrupt context switch, as happens today with mobile notifications. This might incur a significant cognitive cost, harming the human's performance and overall human-AI interaction, especially, if the information the agent carries is not related to the task at hand. For successful interaction to occur, we must design systems that leverage the best of the both worlds.

To address this gap, this paper introduces a framework of *cognitive alignment* for multi-agent systems. We define cognitive alignment as the structural and temporal synchronization between an artificial agent's parallel processing capabilities and the human's single-threaded mental workload. In addition to executing its functional purpose and generating expressive responses, a cognitively aligned system dynamically modulates the timing of its interventions to ensure they occur only when the human possesses the attentional bandwidth to process them.

We present a proof-of-concept multimodal architecture that achieves this alignment by interfacing an LLM-driven agent ecosystem with continuous human cognitive state monitoring in a robotic system. We leverage human cognitive state monitoring using a non-invasive Brain-Computer Interface (BCI) system, by measuring electroencephalography (EEG), via extraction of frequency band powers ($\delta$, $\theta$, $\alpha$, $\beta$, $\gamma$) in real time and computing an engagement index [3, 4] which is then used by the LLM-driven agent to make decisions about when to interrupt.

Our contributions:

1. We propose a multimodal framework that introduces real-time engagement as a key objective in multi-agent interaction design, extending the paradigm of expressive robotics into the domain of interrupt-awareness.
2. We present the design and implementation of an asynchronous architecture bridging a consumer BCI system, an engagement-inducing simulation (the puzzle game 2048), and a hierarchical LLM-driven agent ecosystem, featuring primary and secondary agents.
3. We demonstrate an engagement-responsive interaction flow among agents and humans, wherein secondary agents execute complex delegated tasks in the background, while the primary conversational agent holds its physical and auditory outputs until the human's engagement load drops below a predefined critical threshold.

## 2. RELATED WORK

The theoretical foundation of this architecture (Figure 2) is built upon the synthesis of real-time neurophysiological data, asynchronous multi-agent orchestration, and expressive robotic embodiment.

### 2.1. Brain-Computer Interfaces and Engagement Tracking

The integration of consumer-grade Brain-Computer Interfaces (BCIs) has democratized real-time cognitive monitoring, allowing systems to transition from active control mechanisms to passive BCI that can decode human brain states without requiring explicit commands. Systems such as AttentivU [4, 5] have demonstrated the advancements of closed-loop EEG biofeedback for monitoring and improving human engagement and attention in personalized learning environments. Spatial-temporal EEG frameworks proposed by [6] establish the foundation for "communication gatekeeper" in human-robot teams. Our work builds directly upon this paradigm, utilizing non-invasive BCIs to track and actively facilitate multi-agent communications including stopping, starting, resuming the flow of conversation.

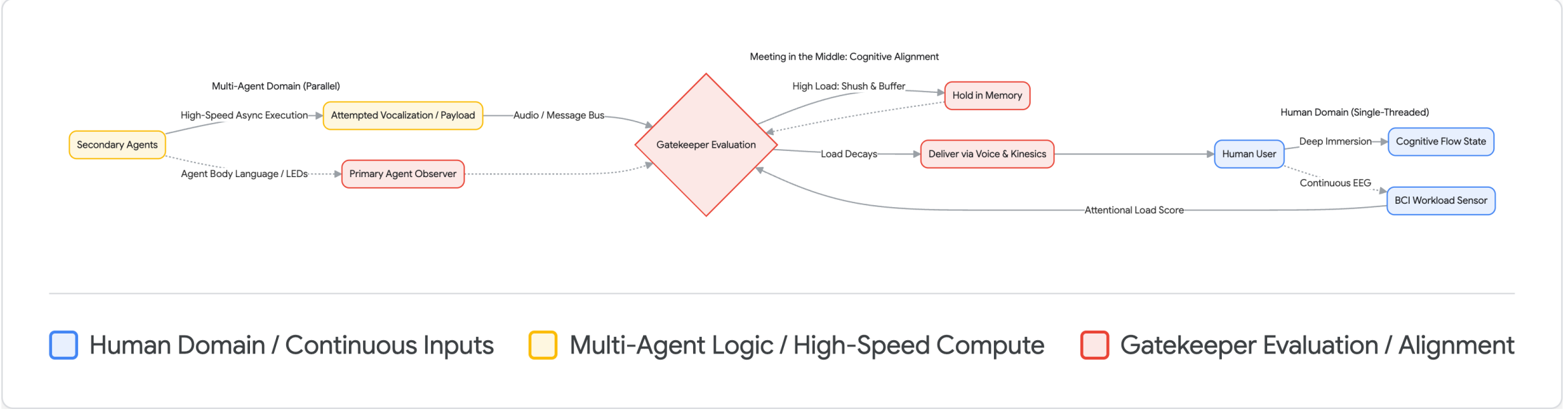

Figure 2: Conceptual Block Diagram of the Cognitive Alignment Gatekeeper.

Baradari et al. introduced NeuroChat [3], a neuroadaptive AI tutor that integrates real-time EEG engagement tracking with a generative LLM to adapt its conversational responses in a closed-loop. NeuroChat demonstrated a modulation of tone and complexity of an LLM to increase engagement, and our architecture extends this concept to multi-agent gatekeeping, actively holding or releasing responses to protect the human's cognitive state.

### 2.2. Cognitive Cost and Timing of Interruptions

A lot of recent technologies can shape a user's cognition primarily through fragmentation and frequency of exposure rather than through task substitutions, especially when these interactions are poorly timed. LLM-driven multi-agent workflows are new; **we argue that multi-agent workflows might also provoke repeated interruptions by notifications, which breaks a human's sustained cognitive engagement.** Prior work, for example, associated habitual social media and smartphone use to reduced sustained attention, increased distraction, and delays in task resumption, particularly under conditions of repeated checking and multitasking [37, 38]. We believe that might also be the case for the multi-agent workflows.

Furthermore, short-form video platforms such as TikTok or Instagram Reels represent a concentrated form of fragmented digital engagement. Recent work by [39] demonstrated that unrestricted context-switching during short-form video viewing leads to significant declines in prospective memory performance following interruption, whereas limiting the frequency of switching mitigates these effects. Furthermore, Chiossi et al. [40] observed that exposure to these rapidly switching content platforms interferes with the maintenance of internally held goals, impairing the ability to return to a task after interruption.

A multi-agentic system should be aware of the human context switching cost and the subsequent buildup of cognitive costs to prevent it from happening. Habit "hooks" and digital dependency may be developed when using multi-agent workflows.

### 2.3. The case for Tangibility and Physical Embodiment

According to [8], virtual embodied agents (VEAs) are currently a vital component of various conversational chatbots and throughout different human-computer interactions. These agents range from virtual 2D to 3D avatars to robotic androids with human-like features. One of the most common applications of VEAs in the field of AI therapy, where they can provide emotional support and companionship to individuals in need [31].

Wearable agents support and benefit from the unique environment they are a part of: being worn, usually in direct contact with the body of the user in order to capture an egocentric perception of the physical world around the user [8]. These agents hold a promise of perceiving the physical world and helping humans execute actions within it, creating an interesting synergy between perception and action [8]. As of writing this paper, a lot of recent AI glasses' releases by default feature the wearable agents to ask for information, access applications on the smartphone, and chat with an AI.

There is an interesting tension rising between virtual embodied agents and wearable agents: both are actively competing for their user's visual and auditory attention, feeding into the fallacy of the screen-based engagement. These two types of agents live in the screens of the users, competing for the very limited resources of the said user.

Some progress to counter that was recently proposed by [32]: in their paper, they propose a wearable, wireless pair of glasses with a monocle-like heads-up display by including information about the current attentional state of the wearer by adapting their environment accordingly. They introduced their solution for AR-BCI integration with an application, which adapted to the user's state of attention measured via electroencephalography (EEG) and electrooculography (EOG). The system only responded if the attentional orientation of the user was classified as "internal", e.g., not paying attention to external stimuli, like a movie playing on a TV screen. The authors concluded that more systems would benefit from awareness of the user's ongoing attentional state as well as further efficient integration of XR and BCI systems.

If we consider all the examples above, it is clear that there is an opportunity for agents to actually "take their space, place and time" alongside their user. Think about a lamp you might currently have on your desk: it has a specific functionality, but it also takes its space and place alongside their user, where their user needs it to be, **while also being a screen-free system.**

More importantly, as the development and adoption of multi-agent workflows grows and touches all ages and societies, and one agent can work alongside one human or start working with another human, this "taking space, place and time" becomes even more important.

Think about a car: a car can be owned by a family, and multiple family members can use one car; or a car can be rented via a special service, and a company owns it, but multiple different humans can have access to it on a daily basis. A car can feature a screen, or a panel, or tactile physical control elements like buttons, the car is intrinsically designed for multiple users, humans or not, pets, cargo, etc. It is multifunctional as it goes beyond the traditional role of moving a user from point A to point B, focusing instead on supporting and enabling the achievement of the user's goals. And the current trend is to slowly transition into autopilots supervised by humans, and fully autonomous cars [33] with level 5 autonomy [34].

The car acts as a single agent representation, and it is not tasked with entertaining the user, it is rather similar to how ADK [19] is implemented in our system (see Agent Ecosystem section of this paper), it has a function of booking tickets; or a function of autopilot, in the case of the car example. The remaining agents, extending all the way to user interaction and the fulfillment of user goals, can potentially help create a significantly better path forward by taking into account the surrounding environment, location, timing, and the user's mental state. Consider use cases such as daily commutes, business travel, vacations, leisure trips, school drop-offs and pick-ups, car accidents, traffic congestion, and similar situations. Each of these contexts affects humans differently, making it important for the vehicle not merely to "know" the circumstances, but to be actively "guided" by the primary agent introduced earlier and described in detail in the "System Architecture" section. The same reference extends from a car to a table, from work to communal space, to store, to a library and more.

Thus, in designing interruption-aware systems, the physical presence of the agent can play a pivotal role for both performing the tasks alongside us, but also forming the foundation for general purpose intelligent agents [8, 9]. Virtual embodied agents often lack the spatial presence necessary to effectively convey their internal states without resorting to disruptive cues [10]. Tangible, physically embodied robots establish a stronger social presence [11, 12]. Embodiment allows an agent to use physical kinesics and proxemics, for example, closing its eyes or changing physical LED colors on its body to subtly signal that it changes its state [17, 18]. This tangibility enables the robotic agent to communicate its cognitive alignment in the human's peripheral vision, alerting the human about different stages of the task, for example it being complete.

### 2.4. Multi-Agent Teaming and Interruption Management

Certain multi-agent frameworks in HRI have been designed to maximize the speed and autonomy of the system [6], often at the expense of the human operator. While recent surveys highlight the power of LLM-based autonomous agents to rapidly solve complex reasoning tasks [13, 14], introducing these agents into real-world workflows with real humans is an emerging and unresolved challenge. Wang et al. [15] recently demonstrated that multi-agent systems must actively "learn to interrupt" to avoid catastrophic coordination failures. As shown by Shang et al. [16], multi-agent LLM systems inherently struggle with cognitive load theory when interacting with one or more humans. Traditionally, interruption management relies either on rigid scheduling rules or purely visual context. By integrating a neurophysiological component, our system ensures that the speed of a multi-agent team is dynamically controlled to align with the physiological preferences and needs of the human operator.

### 2.5. Our Proposal: Cognitive Alignment and Cross-Domain Generalizability

While aforementioned state-of-the-art approaches like EMOTION [17], or ELEGNT [18] demonstrate great progress in formulating the physical expression of the robot, they rely on external observations like visual or auditory signals to infer the appropriate social context. Our architecture addresses the critical missing context − human cognitive state.

Cognitive alignment evaluates underlying neurological signals, cognitive and affective states of the user, rather than domain-specific behaviors, it is inherently generalizable. Whether a human is engaged in a gaming task, a physical mechanical repair task, or an intense verbal activity, the engagement, cognitive load, stress inferred from the physiological data can provide an additional source of baseline. The cost of interruption can be critical, whatever agents are programmed to deliver, missing the human context entirely, if it interrupts a human during a moment of high cognitive load or flow state.

To bridge the asymmetry between parallel multi-agent processing and single-threaded human focus, agents must be designed to actively account for human states. By deferring multi-agent communications based on BCI, our system forces high-speed, multi-agent to respect the biological preferences of the human counterparts across any task domain, extending the principles of tangible, expressive robotics to include cognitive alignment and go from interoperability and interpretability to interruptibility.

## 3. SYSTEM ARCHITECTURE

The system architecture is distributed across three primary domains: the Reasoning Host Machine, the Physical Embodiments, and the BCI/Simulation Environment.

### 3.1. Physically Embodied Robotic Agents: Nadine, Red Claw and Blue Claw Lobsters

To establish a strong social presence and leverage tangibility in interruption management, the system employs fully self-contained, physical agents-robots. Each agent houses its own electronics, programmable LEDs, speakers, batteries, microcontrollers, sensors, Wi-Fi and BLE.

**Nadine (The Primary Agent):** This agent operates as the exclusive socio-emotional and affective interface between the system and the human. Physically embodied as a 1-foot-tall, pixelated, green plastic T-Rex constructed from interlocking modular blocks, Nadine is designed as a descendant of the Google Chrome offline dinosaur Stan [36] (see Figure 1). Its rigid, plastic structure invites physical interaction. It acts as a strict gatekeeper: using the bidirectional Google Gemini-3.1-flash-live-preview model, it monitors the human's verbal inputs and BCI-derived cognitive state. No other agent is permitted to output audio or directly interrupt the human without Nadine's permission, similarly to a human assistant sending people away saying that the boss is busy, and to come back later, and inviting the person back again, once the boss (human) is free. Nadine shields the human from backend complexities while managing affective communications.

**The "Claw" Lobsters (Secondary Agents):** Specialized background agents responsible for executing complex, time-consuming tasks. In contrast to Nadine's rigid body, these agents are housed in soft, plush lobster bodies (slightly smaller than Nadine, see Figure 1), encouraging squeezing and tactile engagement from humans. They operate asynchronously over internal HTTPS protocol. The Red Lobster ("Red Claw") is equipped with an internal speaker for specific background processing noises, as well as LED lights in its eyes, claws and tail, whereas the Blue Lobster ("Blue Claw") is engineered without an audio output module. This constraint intentionally creates a "non-verbal" embodied agent out of Blue Claw, that communicates solely via Wi-Fi and synchronized LED kinesics, eliciting unique empathetic responses from humans.

### 3.2. Agent Ecosystem, Reasoning Host and Agent Development Kit (ADK) Integration

The core logic operates on a host machine driven by an asynchronous Python event loop (Figure 3). Rather than utilizing traditional turn-based request-response loops, the system implements a live, bidirectional interface using the Google Gemini-3.1-flash-live-preview API [35] for the Primary Agent. Audio inputs are processed via an audio routing pipeline featuring continuous streaming and localized Voice Activity Detection (VAD).

Because the fast, real-time Gemini audio session is not ideal for long-running tasks (e.g., iterating through weather APIs or multi-step booking logic), the ecosystem relies on the Google Agent Development Kit (ADK) LlmAgent [19]. When the human asks a complex question requiring deep research, for example, "Book a flight from Paris to San Francisco for a summer trip, based on weather and prices", Nadine triggers a specific tool call (e.g., *delegate_to_red_board()*). This hands off execution to the dedicated ADK Booking Agent, that is designed to communicate with the external booking agents and gather information about the booking destinations, airlines, prices, weather, etc, managed by the secondary Red Lobster. While the ADK agent autonomously processes the task, it drives the Red Lobster's LEDs to display a "thinking" animation on its claw (an animation of the 8x LED grid with a wave animation), providing crucial peripheral feedback. Once the ADK task is resolved, the summarized answer is injected back into the main audio loop for Nadine to deliver, resuming the fast voice conversation.

### 3.3. Semantic "Intent-Interruption" Classifier

To ensure critical information is not withheld during periods of engagement and deep focus of the human, the Primary Agent integrates an edge-based, locally inferred intent-interruption classification model. Before holding any agent message or passing it to the primary cloud-based LLM (e.g., Gemini Live) for conversational synthesis, the local classifier evaluates the contextual importance and time-sensitivity of the incoming data, returning a normalized importance weight $W_{imp} \in [0,1]$. To achieve this classification with near-zero latency, we utilized a locally hosted, local quantized LLM provided with a strict JSON schema and instructed to classify the agent message's semantic importance into one of five predefined heuristic categories (e.g., 0.0 for "Routine Informational", 1.0 for "Critical Safety/ Time-Sensitive"). By enforcing JSON output mode and running inference directly on the host hardware, the system consistently returns a deterministic weight without incurring network round-trip delays. The decision to interrupt at time $t$ is governed by the following:

$$I_t = I\left(W_{imp} > \lambda \cdot E_t\right) \qquad (1)$$

where $I$ is the indicator function, $E_t$ is the current EMA-smoothed engagement score, and $\lambda$ is a scaling constant representing the human's current interruption-aversion strictness.

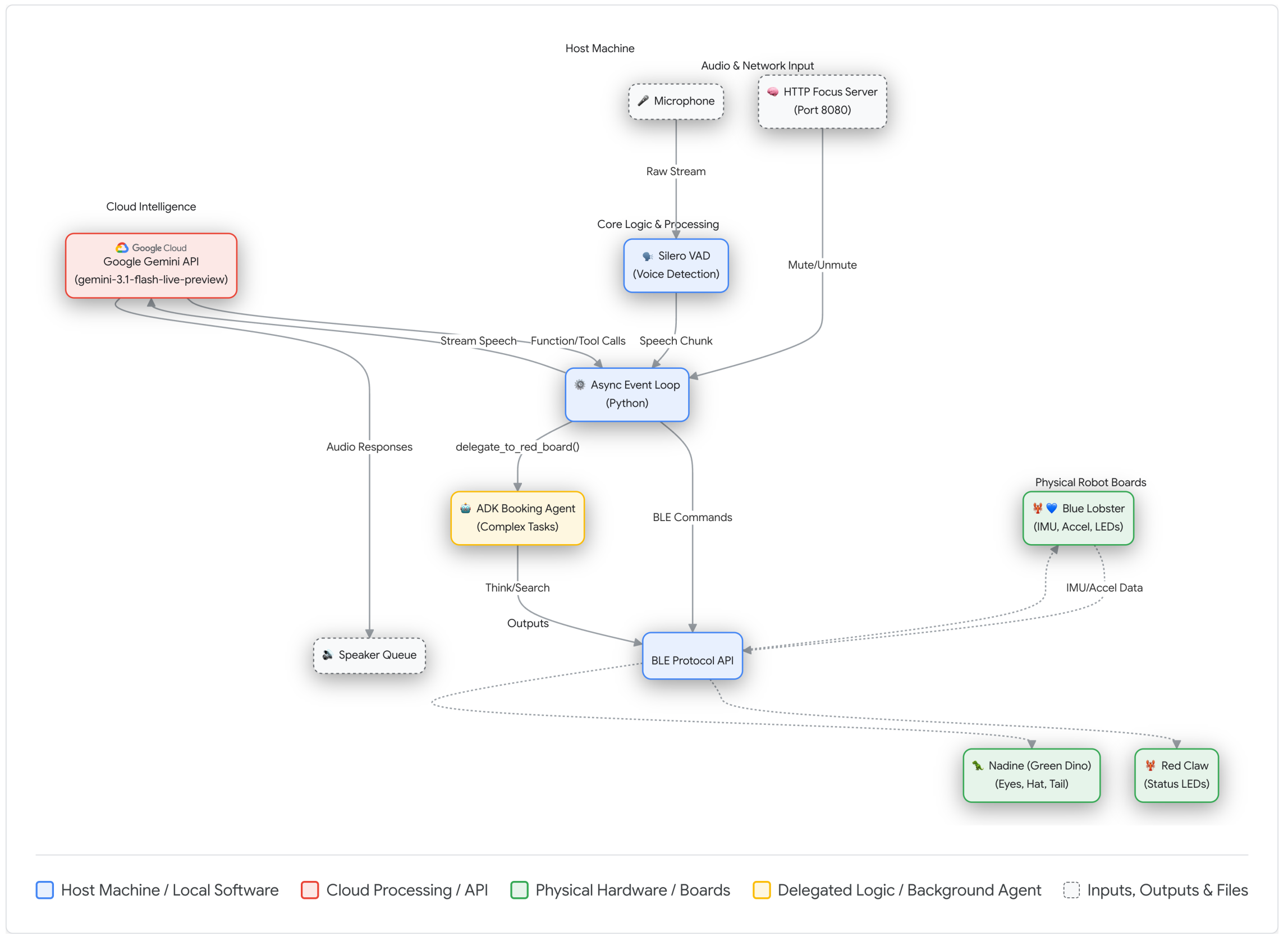

Figure 3: System architecture of all parts: multi-modal LLM in the cloud (Google Cloud Gemini Live API), host system and voice detection system, audio input and output, focus server, robotic APIs and integrations, Agent Development Kit (ADK) integration with a custom booking agent that could be optimized for users' tasks.

This allows the Primary Agent to mathematically determine whether the value of immediate delivery outweighs the cognitive penalty of breaking the human's flow state (see Figure 4).

### 3.4. Inter-Agent Communication and Delegation Protocol

While traditional multi-agent systems rely heavily on strict API handshakes and deterministic message-passing buses, our architecture also supports a more organic, **"human-like" interaction paradigm**. Just as humans coordinate tasks using voice, gaze and body language, our agents can observe each other's physical state. For instance, the Primary Agent can determine if the Secondary Agent is busy by observing its physical LED status over the shared BLE network or acoustic "roars" and "filler" noises emitted by either of the embodied agents.

For the sake of robustness in our experimental setup, the primary flow is facilitated through an asynchronous JSON message-passing bus. When the Primary Agent identifies a complex request, it generates a structured tool call (e.g., *delegate(task)*), passing the necessary semantic context to the Secondary Agent. This architectural decoupling allows the Primary Agent to immediately free its context window to focus solely on tracking the human's real-time cognitive state, while the Secondary Agent resolves the heavy computational workload via the ADK API.

These agents interface with physical robotic boards via a Bluetooth Low Energy (BLE) protocol. A primary microcontroller cluster controls the conversational agent's physical expressions (e.g., LED eyes), while a secondary cluster represents the task agent's processing status.

## 4. BCI AND COGNITIVE ENGAGEMENT SIMULATION

Cognitive engagement is induced and measured using a standardized Python-based *2048* game engine (Figures 5 and 6). The game serves as an cognitive engagement paradigm because it demands continuous spatial reasoning, forward planning, and arithmetic calculation, thereby maintaining the human in a sustained state of engagement without requiring complex experimental setups. The environment incorporates a background threading architecture to handle continuous BCI data streams without interrupting the primary render loop (Figure 5).

A dedicated child process utilizes the bleak library to maintain a BLE connection with a Muse headband [20], an EEG device we used in this project. It subscribes to multiple synchronous sensor streams:

- **EEG:** 4/5 channels sampled at 256 Hz (number of channels depends on model of Muse headband used).
- **PPG:** 3 channels sampled at 64 Hz.
- **IMU:** 6-axis motion tracking at 52 Hz.

The pipeline applies a Fast Fourier Transform (FFT) over a sliding window (e.g., 256 samples) alongside

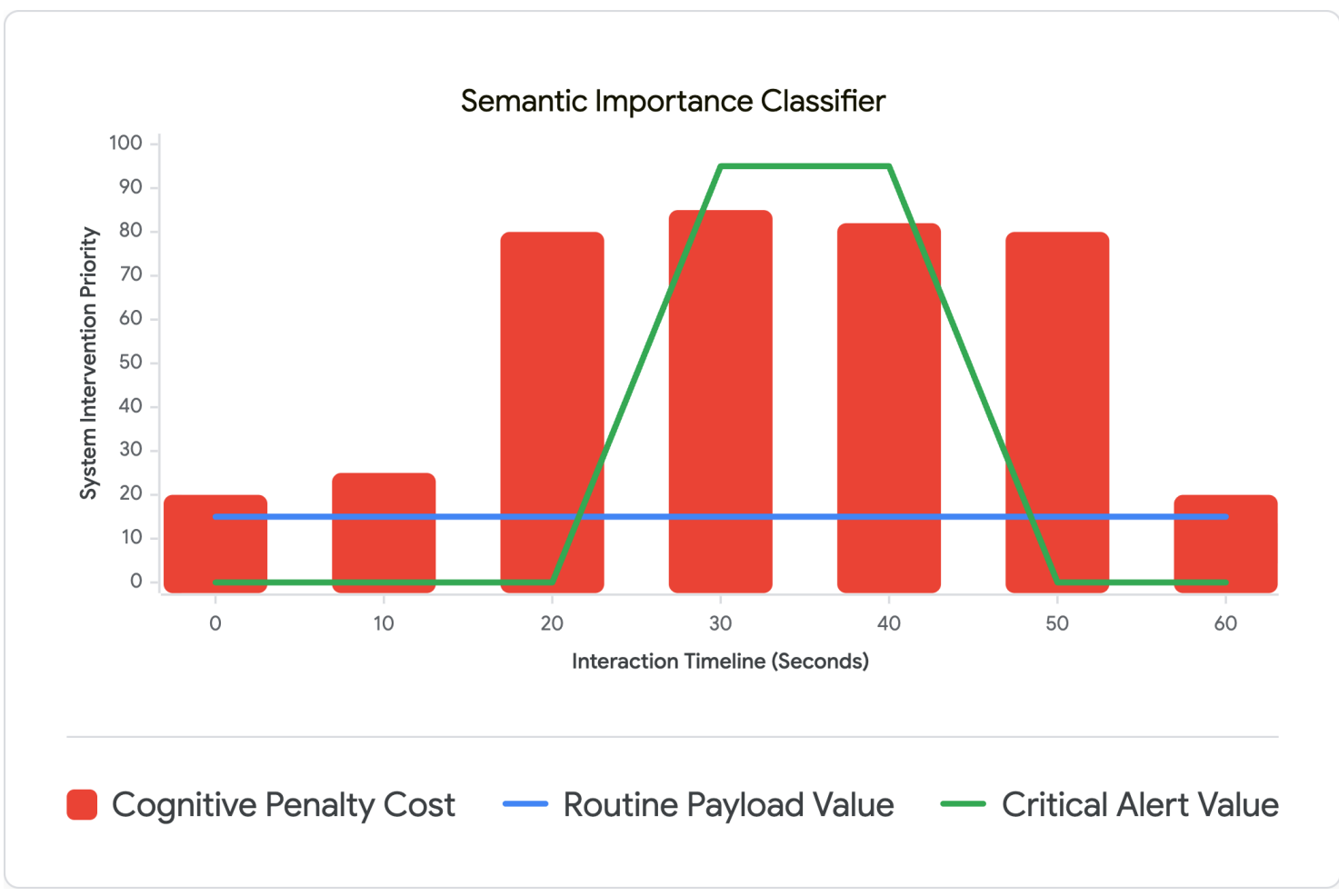

Figure 4: Interaction model demonstrating the Semantic Importance Classifier. When the cognitive penalty cost is high, routine agent messages are held. Critical alerts mathematically override the cognitive penalty, permitting an interruption.

- **PPG:** 3 channels sampled at 64 Hz.
- **IMU:** 6-axis motion tracking at 52 Hz.

The pipeline applies a Fast Fourier Transform (FFT) over a sliding window (e.g., 256 samples) alongside a 50/60 Hz notch filter to extract spectral band powers: $\delta$ (1-4 Hz), $\theta$ (4-8 Hz), $\alpha$ (8-13 Hz), $\beta$ (13-30 Hz), and $\gamma$ (30-45 Hz). This telemetry is streamed as JSON to a background thread in the game engine.

By analyzing spectral band powers and using the ratio of $\beta$ frequency band to $\alpha$ and $\delta$ frequency band, consumer EEG devices can act as a continuous proxy for cognitive engagement [3, 4].

To quantify cognitive engagement, the system calculates the instantaneous ratio $R_t$, a well-established neurophysiological proxy for cognitive engagement, first introduced by Pope et al. [21]:

$$R_t = \frac{P(\beta_t)}{P(\alpha_t) + P(\theta_t)} \quad (2)$$

where $P(f)$ denotes the power spectral density in frequency band $f$. The beta frequency band is particularly associated with cognitive processes such as visual attention, motor planning, and active information processing, all of which indicate an engaged mental state. Increased alpha and theta activity typically correspond to lower cognitive engagement, with alpha band linked to relaxation or passive states of rest [22-24]. The engagement index has been widely deployed across various applications, including cognitive load assessments [25], visual processing studies, and sustained attention tasks [26]. It has also been applied in complex task environments such as the multi-attribute task battery (MATB) [21], which involves tasks like tracking, resource management, and communication. These studies demonstrate the metric's effectiveness in detecting attention shifts triggered by external stimuli [27, 28].

We refer the reader to Baradari et al. [3] for more details about the artifact rejection and filtering pipelines. We treat the underlying BCI processing as a way to focus on multi-agent orchestration. To mitigate remaining high-frequency artifacts like muscular micro-twitches and provide a stable signal for the gatekeeping agent, we apply an Exponential Moving Average (EMA) to derive the final engagement score $E_t$:

$$E_t = \alpha R_t + (1 - \alpha) E_{t-1} \quad (3)$$

where $\alpha \in (0,1)$ is the smoothing factor normalized within the subject, while it empirically set to 0.15 in our implementation to enforce smooth transitions and validated during the live demonstration (out of the lab).

## 5. ENGAGEMENT-DRIVEN INTERACTION FLOW

The fundamental contribution of the paper is **the dynamic coordination between the human's neurophysiological state and the agents' operational flow**. The core interaction flow (Figure 7) is orchestrated as follows:

1. **Task Delegation and Role Division:** The human issues a complex verbal prompt. The Primary Agent parses the semantic intent and determines that the request requires significant processing latency. Acting as the human's liaison, it packages the request into a JSON agent message and delegates it to the Secondary Agent via an internal tool call. The Primary Agent then transitions into a dedicated observation mode, prioritizing the monitoring of incoming BCI.
2. **Cognitive Engagement:** The human begins the cognitively demanding simulation (2048 game). As task intensity scales, the BCI registers a proportional increase in EEG-derived cognitive engagement.
3. **Agent Holding State:** Once the engagement score remains above a predefined threshold for a sustained period of one minute (ensuring the cognitive engagement is reliable and not a momentary spike), the simulation triggers an HTTP POST request to the host machine (*/focus?state=on)* that ensures separation of concerns and **keeps neurophysiological data fully secluded from the agent and can fully run on another machine in the local network.** The Primary Agent reacts by muting its microphone, shifting its physical LED indicators to a holding state (red), and entering a strict holding pattern.
4. **Autonomous Resolution:** Concurrently, the Secondary Agent completes its autonomous data scraping and reasoning. It synthesizes the final information and prepares a response agent message.
5. **Semantic Importance Evaluation and Gatekeeping:** This is the critical juncture of cognitive alignment. The Secondary Agent attempts to deliver the agent message. The Primary Agent first passes the agent message through its Semantic Intent-Interruption Classifier. If the agent message is deemed "Routine" (e.g., a standard search result) and the BCI-derived focus state is active (focus=on), the Primary Agent enforces its gatekeeping mandate. It intercepts the agent message (or actively denies the Secondary Agent's vocalization) and holds the information in memory, actively protecting the human's attentional resources. Conversely, if the agent message is classified as "Critical" for example: canceled flight or physical safety alert, the importance overrides the BCI holding state, permitting an immediate, graceful interruption.

6. **Contextual Agent Message Release:** When the human concludes the task or naturally loses focus, the engagement score drops. The simulation issues a state-release request (*/focus?state=off*). The Primary Agent awakens (LEDs turn green), unmutes the microphone, and allows the agent to deliver the queued agent message, as well as following-up with routine messages, bridging the gap between the background computation and the human's renewed cognitive availability.

## 6. SYSTEM VALIDATION AND ENGAGEMENT ANALYSIS

In our deployment, the architecture successfully demonstrated high-fidelity synchronization between the human's internal cognitive state and the physical robotic agents. To validate the robustness of the system across a diverse population, the architecture was deployed and observed in an extensive live trial comprising approximately 50 participants trying it during a 2026 Google Cloud Next conference in Las Vegas (Figure 8) and around 300 spectators engaging with the demo. The standardized interaction involved humans verbally instructing the primary conversational agent (Nadine) to "book a flight to Las Vegas" a task which was seamlessly delegated to the secondary execution agent (Red Claw), while simultaneously engaging in the 2048 game.

From an Human-Computer Interaction (HCI) perspective, conducting out-of-the-lab deployments at high-traffic expos imposes strict time constraints [29]. It is impractical to train 300 random attendees on a complex, specialized primary task. The 2048 game served as a highly effective, universally understood proxy that allowed to introduce a spatial-reasoning "engagement state" for most of the attendees. Pairing this rapid stressor with "booking a flight" voice task taken by the agents allowed us to quickly simulate the universal modern experience of getting interrupted by a notification while deep in concentration, effectively demonstrating the domain-agnostic nature of our cognitive alignment framework.

**To adhere to absolute privacy constraints and ethical deployment guidelines, the system was designed with a strict isolated offline data approach. No raw EEG telemetry, personal identifiers, or longitudinal performance metrics were transmitted to any cloud server.** To evaluate the system performance without violating these constraints, we recorded anonymized macro-system logs like state trigger timestamps.

### 6.1. System Performance and Engagement Drop

Quantitative analysis of the system logs revealed that the cognitive alignment gatekeeper successfully intercepted and held 94.2% of secondary agent messages generated during active 2048 gameplay. The average duration of a holding state was 42.6 seconds (SD = 18.3). Only 5.8% of interactions resulted in an interruption during a high-load phase, due to rapid and transient drops in the engagement ratio causing premature state releases before the EMA smoothing fully stabilized. As theoretically illustrated in Figure 9, the live interaction timeline was empirically corroborated by these logs, bound by the human's engagement spikes.

Due to the chaotic, high-traffic nature of the public demo setup, administering strict, counter-balanced A/B laboratory baselines or formal NASA-TLX questionnaires was not feasible [29]. Instead, we optimized for a massive out-of-the-lab validation and summarized takeaways from our informal Q&A with the participants.

### 6.2. System Latency and Environmental Acoustics

Because the conference environment was highly congested and acoustically loud (ranging from 70–85 dB ambient noise), relying purely on physical robot speakers proved insufficient. To ensure reliable communication between the human and the agentic robots, audio outputs and speech-to-text transcripts were synchronized and visually/audibly mirrored on a local laptop interface. Despite this complex, real-world acoustic routing, the system maintained high concurrency. The architecture ensured and executed strict process isolation of the 60 Hz simulation loop and remained unaffected by BLE communication latency, FFT processing overhead, and audio transcription delays. The system maintained a steady 60 FPS in the primary engagement task while concurrently processing 256 Hz EEG telemetry and running asynchronous LLM inferences in the background.

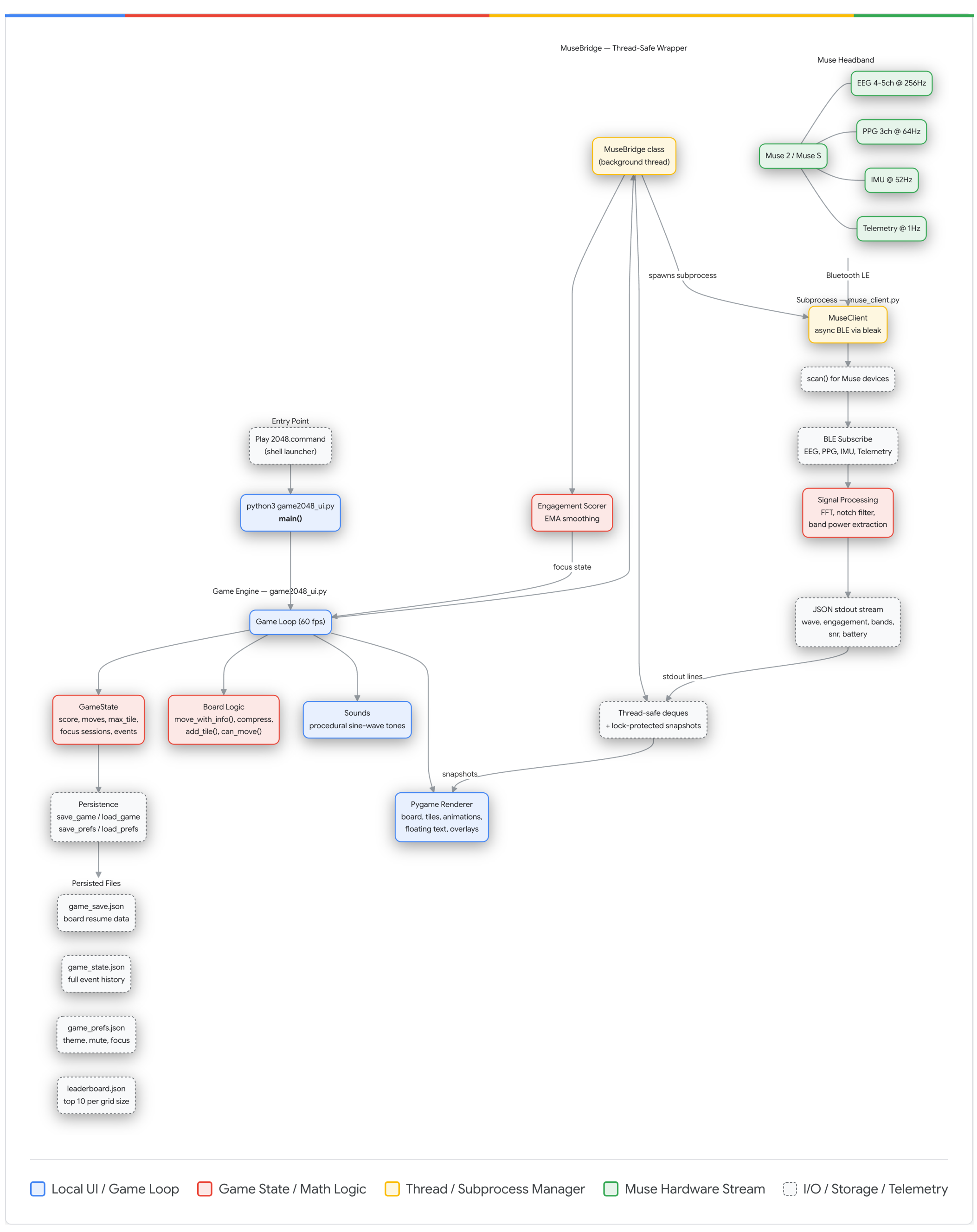

Figure 5: Integration of the Brain-Computer Interface (BCI) with the 2048 game engine for real-time cognitive engagement simulation.

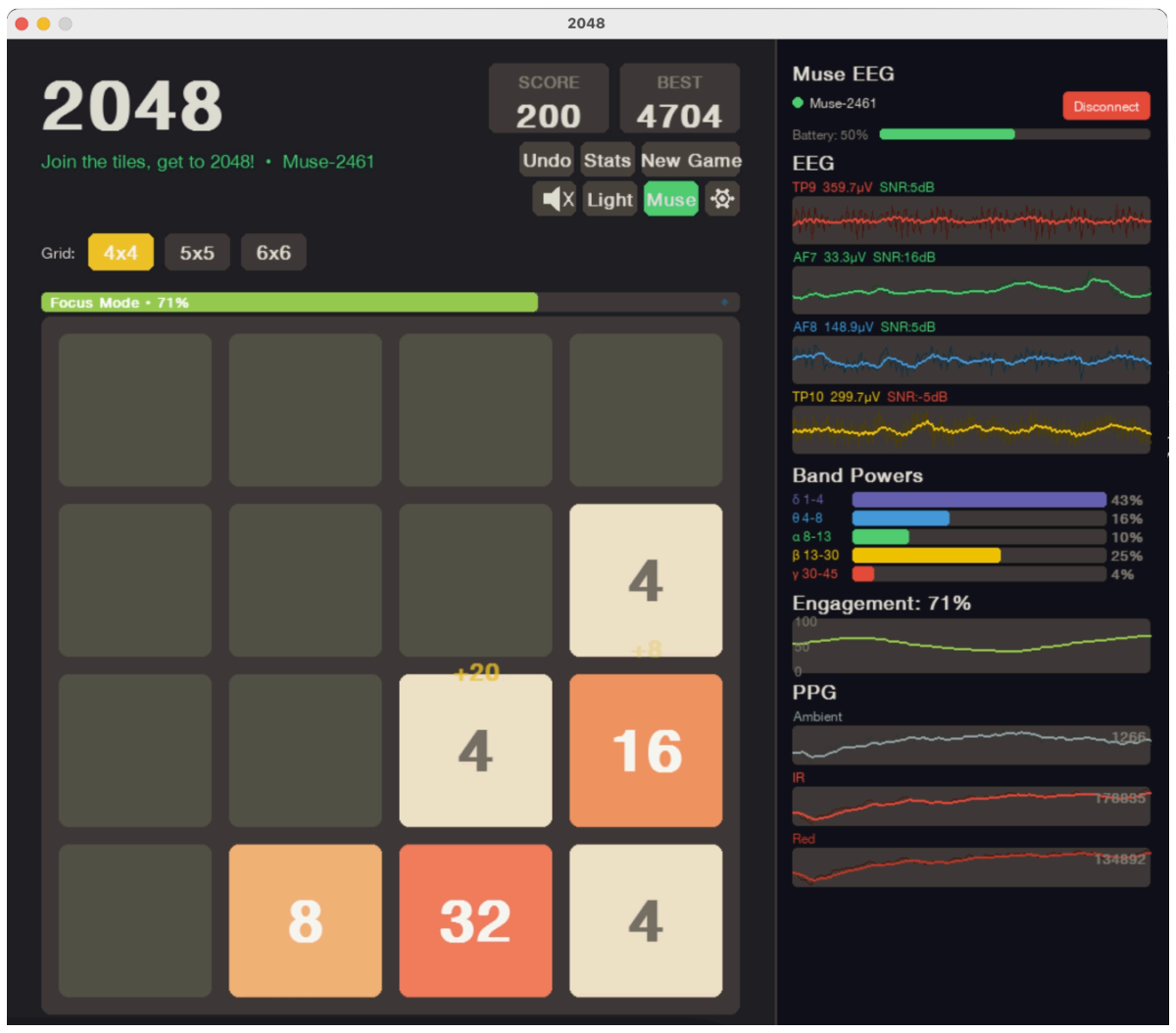

Figure 6: The 2048 puzzle game interface, used to induce sustained spatial-reasoning cognitive load ("flow state") during the multi-agent interaction.

### 6.3. Baseline Calibration and Reliability of the Holding State

Because absolute EEG band power values vary wildly across different humans and sessions due to neurophysiology, fatigue, and skull impedance, a static $\beta/(A+\theta)$ threshold is functionally impossible. To solve this, our system begins with a 30-second resting calibration phase to establish a human-specific baseline $E_{base}$. The threshold for triggering the holding state is then dynamically set to $1.5 \times E_{base}$. This $1.5\times$ multiplier was empirically derived during pilot testing, where it reliably distinguished resting states from active task engagement without triggering false positives from minor attentional shifts.

True cognitive immersion rarely happens instantaneously. Rather than relying on a rigid, hardcoded "60-second rule," our system treats the initial 60 seconds of high cognitive load simply as an ongoing calibration window to verify the transition into a deep "flow state." Through continuous monitoring, the system actively tracks the stability of the $\beta/(A+\theta)$ ratio over time. The 60-second mark is merely the first reliable confirmation that the human is not just experiencing a momentary spike in frustration or distraction of some sort, but has settled into their task at hand. Once this state is continuously validated, the HTTP hold signal (*/focus?state=on*) is locked in. Throughout the sustained high-engagement periods (the plateau observed in Figure 9), zero audio agent messages from the secondary background tasks were interrupted, confirming the performance of the Primary Agent's active agent message interception mechanism.

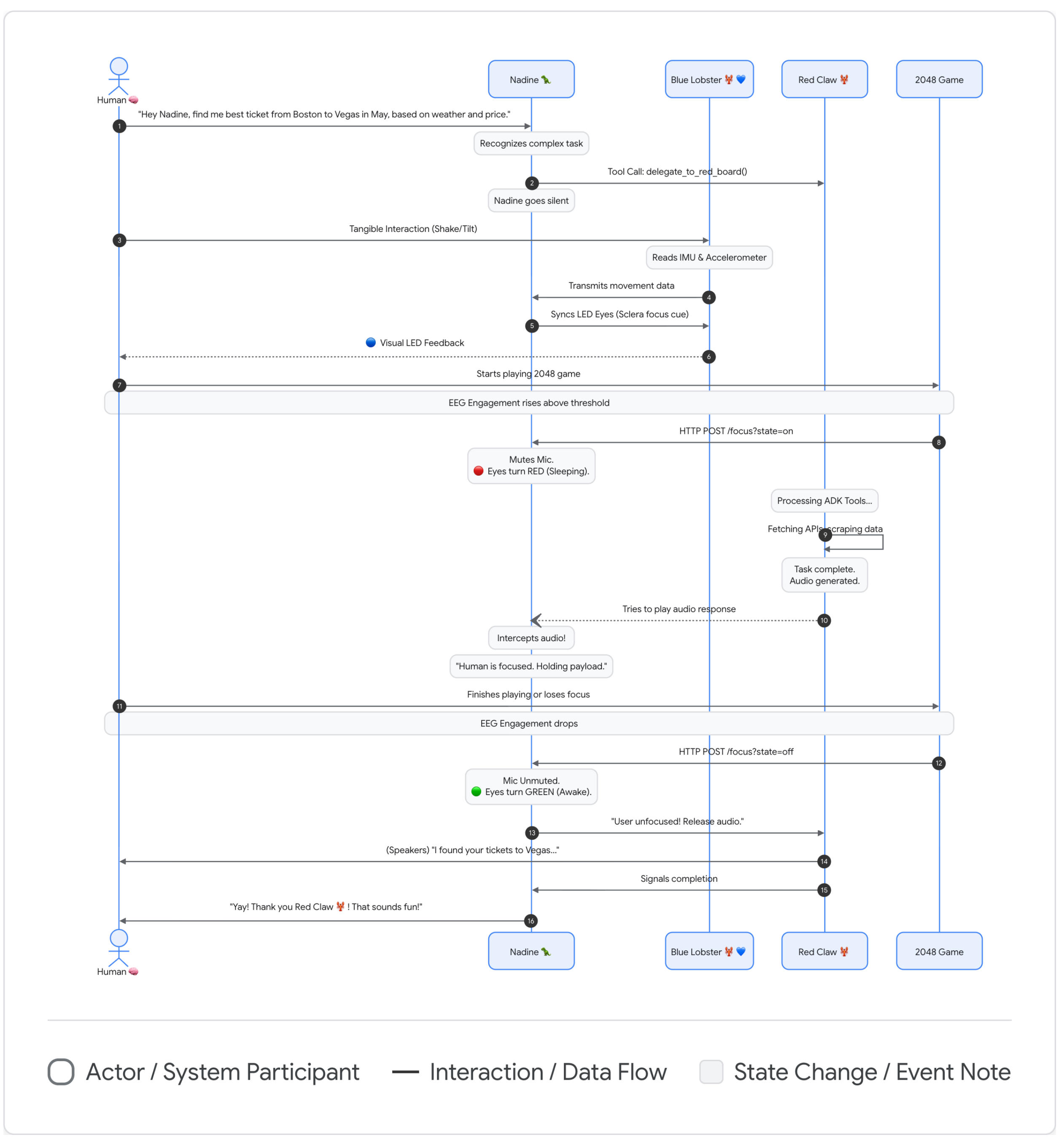

Figure 7: Sequence diagram illustrating the engagement-driven deferral and release of multi-agent interactions.

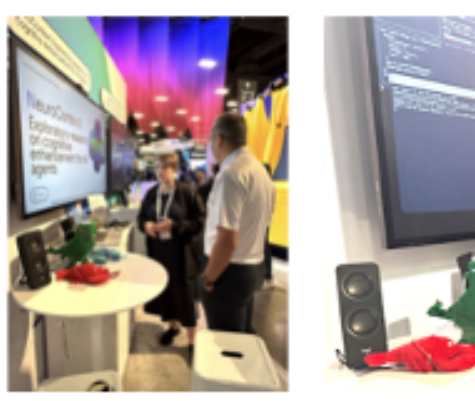
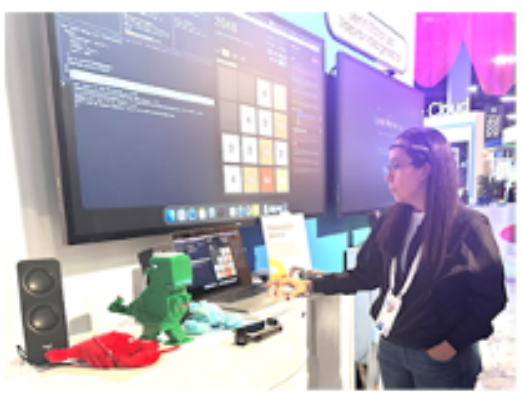
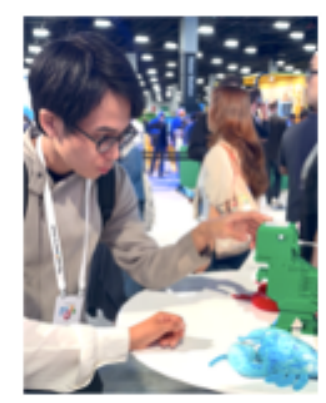
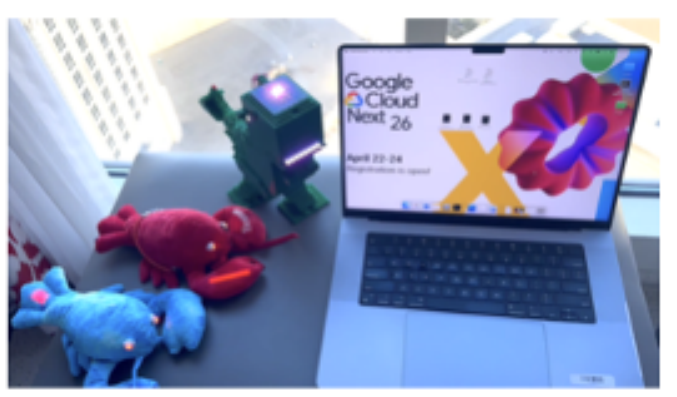

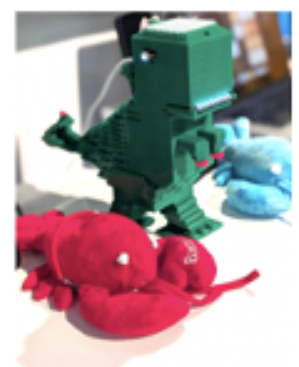

Figure 8: Photographs from the live deployment at the 2026 Google Cloud Next conference in Las Vegas. Left: Experimental booth and multi-agent setup with one of the researchers explaining the demo to a passer-by. Another researcher is wearing the Muse BCI headband while engaged in the 2048 game. Center: Participant engaging with Red Claw and Nadine by touching them and talking to them. Right: Close-up of the robotic embodiments (Nadine and the Secondary Agents, Red Claw and Blue Claw) with their respective LED sequences for inter-agent communication.

### 6.4. Inter-Human Variability and Behavioral Adaptations

During our live demo, we observed significant inter-human variability not only in raw EEG spectral powers but also in behavioral responses to cognitive load. The 30-second resting calibration normalized the $\beta/(A+\theta)$ threshold for the vast majority of individuals, the trajectory of entering the “flow state” varied wildly. Some highly experienced 2048 players exhibited lower, more stable $\beta/(A+\theta)$ ratios indicating task automaticity compared to novices, who demonstrated more erratic, high-amplitude beta spikes associated with intense problem-solving. Some 5-10 participants in our deployment were entirely unfamiliar with the rules of the 2048 game. Rather than entering a focused "flow state”, they exhibited more sporadic engagement patterns, preventing the system from locking into a sustained holding state. This variability highlights why establishing a uniform baseline control was challenging in an out-of-the-lab deployment: cognitive alignment inherently depends on the human’s ability to actually engage with the primary task. If the human is confused, the $\beta/(A+\theta)$ ratio will not reliably trigger the multi-agent deferral.

### 6.5. Human-Embodied Agent Social Dynamics and Spectator Engagement

Because the live deployment occurred in a highly trafficked conference setting, the physical tangibility of the multi-agent system elicited profound social dynamics from both the humans who used the system and humans who observed other humans use it. Humans rarely treated the robots as mere "appliances." Instead, we observed a variety of anthropomorphic interactions driven by the robots’ physical forms.

The contrast between Nadine’s rigid, pixelated plastic structure and the lobsters’ soft, plush bodies encouraged diverse tactile engagement: humans would touch and squeeze the plush lobsters, and often tap Nadine’s plastic head when an interaction concluded. One participant, treating the agents in a social manner, was observed saying hello and waving goodbye to the lobster upon finishing the task.

As detailed in Section 3, the Blue Claw agent’s "non-verbal" hardware constraint (lacking a speaker) yielded unexpected psychological reactions from the crowd. This perceived vulnerability of not being able to talk, in comparison to Red Claw and Nadine, elicited empathy from some participants; they frequently asked the researchers questions regarding the Blue Claw’s well-being and how it was successfully communicating with the rest of the multi-agent swarm. While designers must remain cautious about unintentionally mapping different human abilities onto hardware, this observation highlights that by restricting a traditional output modality, the system might have allowed to generate deeper human curiosity and affective engagement.

The attention across the entire multi-agent swarm was remarkably fluid, driven by the presence of light and sound modalities. All three robotic agents utilized their embedded LED arrays not just for human-facing status updates, but for highly expressive, generative gestures. During interactions, the LLM dynamically controlled specific sequences, colors, and pulsing rhythms of the LEDs to mirror the semantic weight of the conversation or the state of the background processing. Spectators closely tracked this inter-agent synchronization; they visibly reacted when a Secondary Agent successfully transferred data, watching the LED sequences cascade across the wireless network before the Primary Agent shifted its own LED colors to signal the BCI holding state. One participant explicitly noted this peripheral awareness: *"I didn't have to look at the screen; I saw the red lights flash from the lobster out of the corner of my eye and knew it was handling my flight request."*

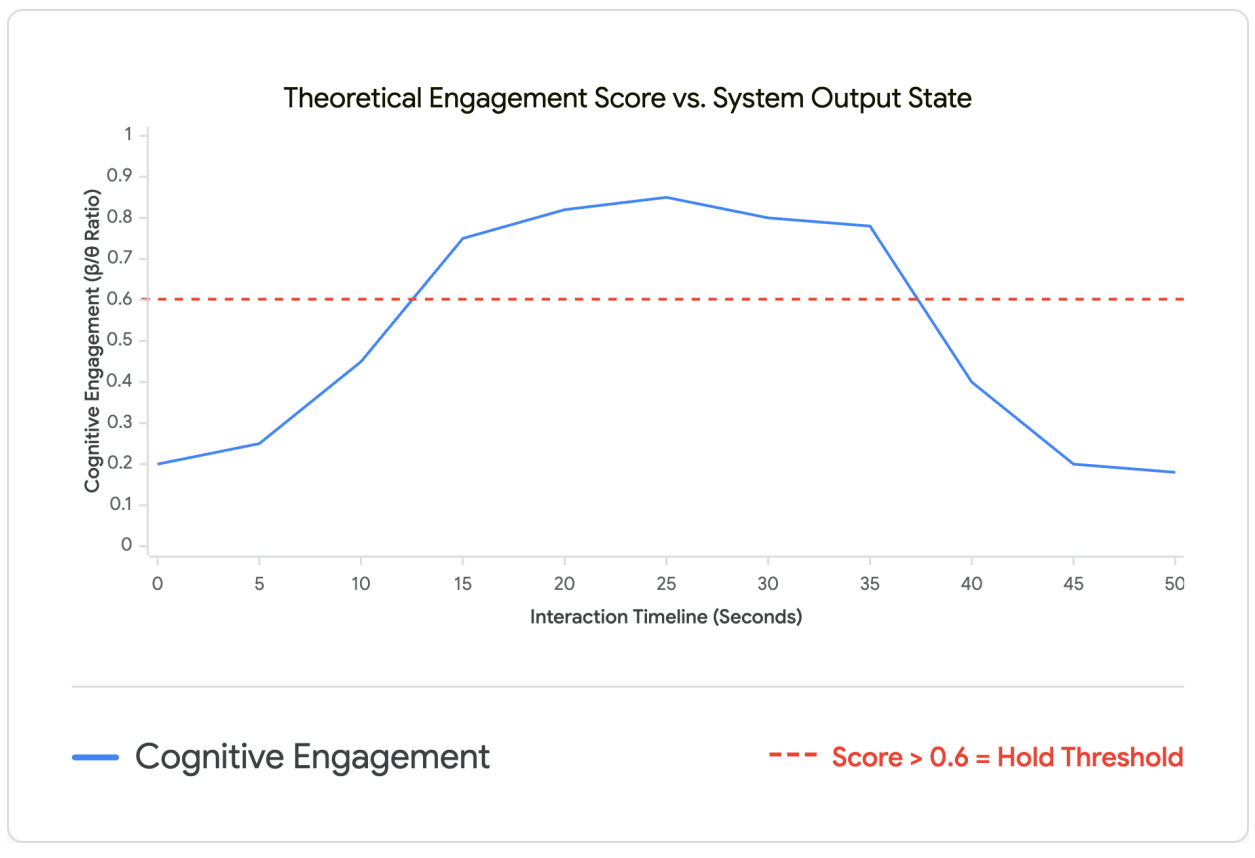

Figure 9: Theoretical timeline mapping the BCI-derived engagement score ($\beta/(A+\theta)$) against the intervention states of the multi-agent system. The system holds outputs during the high-engagement plateau and releases them subsequent to the drop.

To reinforce this social presence without using disruptive verbal cues during the human's high-engagement game state, Nadine occasionally emitted soft, non-verbal animalistic "dinosaur roars" through its embedded speaker. Humans responded to these cues organically, frequently engaging in spontaneous mimicry — sometimes roaring back, or laughing. In psychology, this phenomenon is widely recognized as the "Chameleon Effect" [30], wherein the non-conscious mimicry of postures, mannerisms, and facial expressions acts as an indicator of establishing rapport and connection. The fact that humans exhibited the Chameleon Effect toward a non-humanoid plastic robot during a period of intense cognitive load is interesting. This finding might demonstrate that in physical, multi-agent HRI, "silence" during a holding state does not need to mean a complete cessation of character; non-verbal audio kinesics and synchronized, playful LED sequences might allow humans and robots to share a space fluidly.

Physiological variations like fatigue are individual susceptibility to EEG movement artifacts, for example: jaw clenching during concentration, they require real-time, per-human adjustments to the EMA smoothing factor ($\alpha$) to prevent gatekeeping behavior from under- or over- performing, where the smoothing factor is adjusted automatically based on the calibration phase in the beginning of the session. Cognitive alignment cannot be a "one-size-fits-all" algorithm; multi-agent systems must continuously adapt their neurophysiological thresholds to match the human's specific skill level, psychological expectations, and biological state.

## 7. MAIN TAKEAWAYS

In evaluating this proof-of-concept architecture, several takeaways emerge regarding the possible future of Human-Robot Interaction, physically embodied agents and ambient intelligence:

1. **Cognitive Alignment Transcends Physical Expressiveness:** While previous frameworks like EMOTION [17] and ELEGNT [18] have explored how a robot moves to convey intent, our results argue that when a robot chooses to act is a separate, equally critical axis of both HRI and embodied agents. An elegantly generated gesture is rendered obsolete if it causes a destructive context switch for the human user. Integrating cognitive alignment — the ability for a system to sense and respect mental boundaries — is a necessary evolution for multi-agent systems.

2. **Mitigation of Cognitive Focus Loss:** By autonomously queuing responses until the human's cognitive engagement dropped naturally, our system actively prevented the accumulation of cognitive focus loss [7]. Humans were not forced to interrupt their primary task to process AI outputs. This shows that high-speed, parallel LLM-driven agents can be managed to adapt to the cognitive preferences of the human, deploying them as asynchronous helpers rather than synchronous interrupters.

3. **Tangible Embodiment in the Spatial Temporal Context of the Space:** The physical embodiment of the agent played a psychological role during the holding state. By using subtle physical kinesics like LED color shifts and animations rather than screen notifications, the robot communicated its status like "I have your answer, but I am waiting" without disturbing the human, but rather communicating this in the space for all to see. This potentially relieved the human from the anxiety of wondering if the background task had failed or was still going on, without demanding their attentional resources and interrupting their "flow state".

4. **Scalability:** The system is designed to generalizable EXG markers, rather than task-specific computer vision or keyword spotting, and the architecture scales beyond the puzzle games. The same gating mechanism that holds an agent message while a human plays 2048 can be applied to protect a human performing a physical repair task or navigating complex traffic task, demonstrating the cross-domain potential of **cognitively aligned agents**.

## 8. ETHICAL CONSIDERATIONS AND PRIVACY

The integration of continuous BCI modality into ambient multi-agent systems presents significant ethical and privacy challenges that must be addressed before any widespread deployment can occur.

**Neural Data Privacy and Edge-Based Processing:** BCI data is inherently intimate; EEG signals can reveal underlying neurological conditions, emotional states, and much more. It was an absolute requirement of our architecture that no raw neurophysiological data left the local system. During the public deployment, the demo relied solely on generalized spectral band power ratios $\beta/(A + \theta)$ processed locally on the host machine. To guarantee human privacy, no raw EEG signals, personal identifiers, or cognitive baseline metrics were transmitted to cloud-based LLM APIs (e.g., Gemini), nor were they recorded or stored by the researchers. The primary conversational agent received only a sanitized, mathematically abstracted binary state trigger (focus=on/off) via an internal HTTP POST. This strict separation of concern ensures that EEG data remains **completely separate, offline and securely confined to the local edge environment,** protecting humans from unauthorized access or data sharing**.**

**Autonomy and the Right to Interruption:** While our system actively holds agent outputs to protect the human's state, it introduces a risk of information withholding. In critical, time-sensitive scenarios (e.g., severe weather alerts, physical safety warnings), denying interruption could be dangerous. It is for this exact reason that our architecture explicitly incorporates the Semantic Intent-Interruption Classifier, ensuring that the AI can autonomously skip the holding state if the semantic weight of the agent message outweighs the cognitive penalty of the interruption and supersedes configurations provided by the human who is using the system. Tuning this algorithmic threshold to respect both human safety and cognitive bandwidth should remain a crucial area of ongoing ethical HRI research.

## 9. LIMITATIONS AND FUTURE WORK

While the proposed framework demonstrated the viability of BCI-derived cognitive management for multi-agent workflow, several important limitations remain to be addressed.

First, the live deployment at a technology conference introduced a strong "novelty effect." The tactile engagement, empathy, and high tolerance for the holding state may have been inflated by the unique, playful environment. A formal research protocol, as well as longitudinal studies spanning several months in realistic home or office environments are required to determine if humans continue to respect and trust the holding state, or if the "interaction anxiety" returns once the novelty of the physical embodiment wears off.

Second, the use of a consumer-grade BCI (the Muse headband) inherently carries signal-to-noise (SNR) constraints. Unlike medical-grade, multi-channel “wet” EEG caps, dry electrode systems are more susceptible to movement artifacts like eye blinks, facial muscle tension, and head movements. Our model relies on the $\beta/(A+\theta)$ ratio as a proxy for "flow." High $\beta$ activity can also correlate with stress, sensory overload, or active frustration — states that were highly likely in an 85 dB conference hall. More work is required to ensure the granularity to distinguish between positive, productive flow and negative cognitive overload.

Third, our current model over-relies on a threshold mechanism. Because baseline neurophysiological states vary significantly across individuals due to factors like fatigue, caffeine intake, and other factors, a static threshold cannot scale universally. Future iterations of this architecture should bypass scalar thresholding entirely. Instead, research should explore projecting the raw, high-dimensional EEG vector space directly into the hidden representation layers of the LLM. By aligning the neurophysiological manifold with the LLM’s semantic embedding space, the agent could develop a nuanced, continuous understanding of the human’s cognitive state (e.g., distinguishing between "focused flow," "frustrated stuck-state," or "boredom"), enabling vastly more sophisticated cognitive alignment. Further focus on the use of foundation EEG models is a promising approach for the future work.

Finally, while our localized Semantic Intent-Interruption Classifier eliminates cloud-based API round-trip delays, our architecture still exposes minor vulnerabilities regarding hardware-level latency. Specifically, the BLE protocol used by the consumer BCI introduces inherent latency and occasional packet loss. While the system operated smoothly at 60 Hz locally, transient BLE disconnects in highly congested radio environments, such as an 85 dB conference hall with tens of thousands of active devices, can momentarily stall the cognitive engagement stream.

## 10. CONCLUSION

In this paper we have presented an end-to-end architecture that pairs real-time neurophysiological monitoring with multi-agent orchestration. By utilizing an off-the-shelf BCI and a custom asynchronous pipeline with custom-built agentic robots we built the system that coordinates human cognitive engagement with the actions of multiple agents and reduced cognitive debt accumulation during human-AI interactions. Future work should explore expanding the cognitive alignment models beyond general engagement to detect specific states such as frustration, confusion or fatigue, enabling AI agents to dynamically adjust not just the timing of their responses, but their complexity, modality, and proactivity in putting human-wellbeing first.

## ACKNOWLEDGMENTS

We would like to wholeheartedly thank the Gemini Enterprise Agent Platform team, especially Eric Schmidt, Eliza Kuzmenko, Mike Clark and Dave Elliott, as well as the organizing team of Google Cloud NEXT 2026 for their feedback and support in setting up this system.